\documentclass[10pt,twocolumn,letterpaper]{article}

\usepackage[pagenumbers]{cvpr} 

\definecolor{cvprblue}{rgb}{0.21,0.49,0.74}
\usepackage[pagebackref,breaklinks,colorlinks,allcolors=cvprblue]{hyperref}

\usepackage{microtype}
\usepackage{enumitem}
\usepackage{mathtools}
\usepackage{float}
\usepackage{booktabs}
\usepackage{multirow}
\usepackage{adjustbox}
\usepackage[table]{xcolor}

\newcommand{\G}{\mathcal{G}}
\newcommand{\M}{\mathcal{M}}

\newcommand{\Pool}{\operatorname{Pool}}

\usepackage{indentfirst}
\def\confName{CVPR}
\def\confYear{2027}

\title{Transfer Once, Ask Locally: \\ Compact Audio-Visual Memory for On-Device Question Answering}

\author{%
Qitong Wang$^{1,2}$\thanks{%
Work done during an internship at Dolby Laboratories, Inc. 
Corresponding author: \texttt{wqtwjt@udel.edu}.},
Jihui Jin$^1$,
Divya Kothandaraman$^1$,
Vishnu Raj$^1$,
Christopher Rasmussen$^2$\\
$^1$Dolby Laboratories, Inc.
\qquad
$^2$University of Delaware
}

\begin{document}
\maketitle
\begin{abstract}
Audio-visual question answering on lightweight user devices requires balancing speed, memory usage, privacy, and communication.
One approach is to run audio-visual encoders on a server and a lightweight language decoder on the user device, keeping questions local while offloading media encoding.
However, transferring and processing the full multimodal representation remains costly.
We propose a training-free, query-agnostic token pruning framework that constructs a compact, reusable memory before transmission.
We define video spatio-temporal blocks and aligned audio windows as players and estimate their importance through Monte Carlo Shapley valuation.
We use an embedding coverage objective that combines query-independent utility with cross-modal redundancy calibration and rewards unimodal diversity through marginal coverage gains.
We retain the original token embeddings of high-scoring players under a limited token budget and transfer the resulting memory once for reuse across subsequent local questions.
Experiments on Video-MME, Daily-Omni, and Video-Holmes using video-SALMONN 2+ (3B) and video-SALMONN 2 Pro (4B) show that our method achieves the best or tied-best QA accuracy among the compared compression methods across all evaluated settings.
On Daily-Omni with video-SALMONN 2+ (3B) at 80\% pruning, our method improves accuracy by 12.86--14.62 percentage points over these methods while achieving the lowest device-side LLM prefill and decoding latency, with up to $1.92\times$ decoding speedup at comparable embedding memory usage.
\end{abstract}    
\section{Introduction}
\label{sec:intro}

Interactive media services permit viewers to seek explanations and information about what they are watching.
Second-screen companion applications support this need by providing content-related information on personal devices alongside a shared display~\cite{10.1145/3025453.3025459}.
Conversational question answering helps viewers better understand video content and can further support accessibility, as illustrated by recent interfaces for blind and low-vision viewers~\cite{cheema2026vidscribe}.
For these interactions, answering questions locally can help protect privacy by keeping viewers' questions and responses on their devices.
Apple's Foundation Models framework illustrates this possibility by supporting on-device language processing without sending user inputs to a server~\cite{apple2025foundationframework}.

\begin{figure}[t]
  \centering
  \includegraphics[width=\columnwidth]{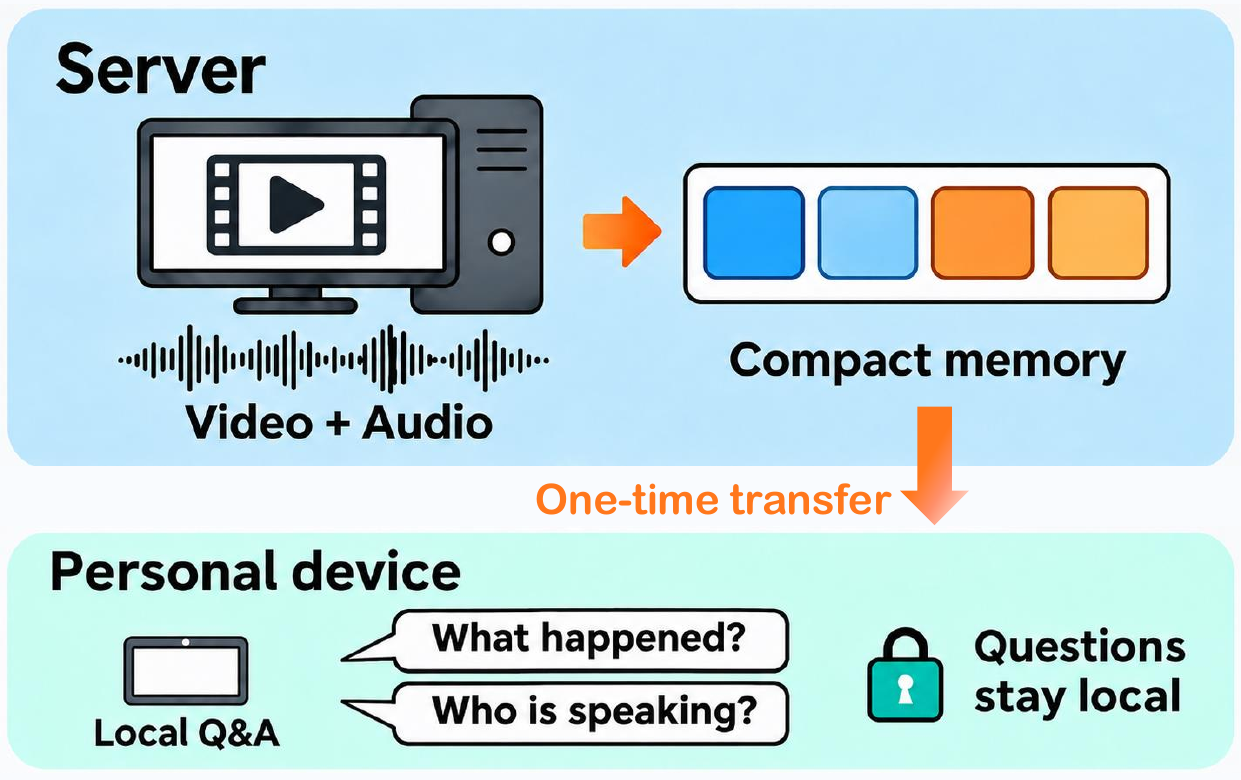}
  \caption{
  A server processes video and audio into a compact memory and transfers it once to a personal device.
  The device reuses this memory to answer multiple questions locally, keeping user queries on the device.
  To construct the compact memory, we propose a novel token selection method. 
  Fig.~\ref{fig:overview} provides an overview of our framework and the proposed method.}
  \label{fig:teaser}
\end{figure}

Multimodal foundation models~\cite{videollava, videollama, videosalmonn2} offer a natural way to answer viewers' questions about audio-visual content.
However, deploying this capability involves a trade-off between privacy and device resources.
On the one hand, processing everything on a server requires transmitting users' questions and potentially personal context beyond their devices, raising privacy concerns and incurring additional communication overhead through repeated question--response exchanges.
On the other hand, running the entire model locally places substantial computational and memory demands on lightweight devices, particularly for long-form video and audio.
Building on prior work on partitioning multimodal models across devices~\cite{11183729}, we consider a collaborative deployment to address these concerns (see Fig.~\ref{fig:teaser}): a server or playback host runs the audio-visual encoders, while each user's device runs a lightweight language decoder.
This division allows the host to encode the media once and the device to reuse the resulting features across questions without sending user queries to the server.
Even with the encoders offloaded, long-form content can produce a large number of multimodal tokens.
Transferring these tokens consumes bandwidth, while storing and processing them increases device memory usage and inference latency, making communication and computation joint considerations~\cite{10.1145/3093337.3037698}.
The central challenge is therefore to reduce the size of the transferred representation while preserving the evidence needed to answer users' questions.

Pruning visual and audio tokens before they enter the device-side LLM offers a way to reduce both communication and local inference costs.
However, several existing approaches rely on information unavailable during server-side memory construction.
Query-guided methods such as PruneVid~\cite{prunevid} require the user's question, whereas our setting keeps questions and personal context on the device.
Pruning inside the device-side LLM, as in FastV~\cite{fastv}, leaves the initial token transfer unchanged and still requires processing the input through the layers preceding pruning.
These approaches therefore cannot directly construct a query-agnostic memory for one-time transfer and reuse across future questions.
Pre-LLM compression methods provide applicable building blocks, but selecting complementary audio-visual evidence remains a key challenge.
OmniZip~\cite{omnizip} uses audio retention to guide video token allocation and cross-modal similarity to consolidate audio anchors.
However, high cross-modal similarity may also indicate duplicated evidence.
When audio and video convey the same information, retaining both may consume budget without adding much coverage.
Whether prioritizing such shared evidence best preserves information for unknown questions therefore remains an open question.

To address these challenges, we propose a training-free, query-agnostic framework that constructs a compact audio-visual memory before transmission to the user device.
We organize video tokens into spatio-temporal blocks and audio tokens into aligned temporal windows, treating each group as a player.
Rather than scoring players in isolation, we evaluate their average marginal contributions to an embedding coverage objective through Monte Carlo Shapley estimation.
The objective combines embedding-based utility with cross-modal redundancy calibration, while encouraging unimodal diversity through coverage.
This formulation assigns less additional credit to already represented content and downweights evidence duplicated across modalities.
The selected groups preserve their original token embeddings, forming a reusable memory that is transferred once.
A lightweight on-device decoder then answers subsequent questions using this memory, reducing communication and local processing costs while keeping user questions on the device.

Our contributions are threefold:
\begin{itemize}
  \item We formulate audio-visual token pruning as query-agnostic memory construction at the server--device interface.
  The compact memory supports one-time transfer and repeated local question answering while keeping user queries on the device.

  \item We develop a training-free framework that treats structured audio-visual token groups as players and estimates their marginal contributions through Monte Carlo Shapley valuation.
  Its embedding coverage objective combines modality-specific utility and cross-modal redundancy calibration, encouraging unimodal diversity and complementary evidence under a restricted token budget.

  \item We evaluate our framework on Video-MME, Daily-Omni, and Video-Holmes using video-SALMONN 2+ (3B) and video-SALMONN 2 Pro (4B), achieving the best or tied-best QA accuracy across all settings.
  On Daily-Omni with video-SALMONN 2+ (3B) at 80\% pruning, our method outperforms competing compression methods by 12.86--14.62 percentage points.
  In this setting, it also achieves the lowest device-side LLM prefill and decoding latency, with up to $1.92\times$ decoding speedup over these methods at comparable embedding memory usage.
\end{itemize}

\section{Related Work}
\label{sec:related}

\begin{figure*}[t]
  \centering
  \includegraphics[width=\textwidth]{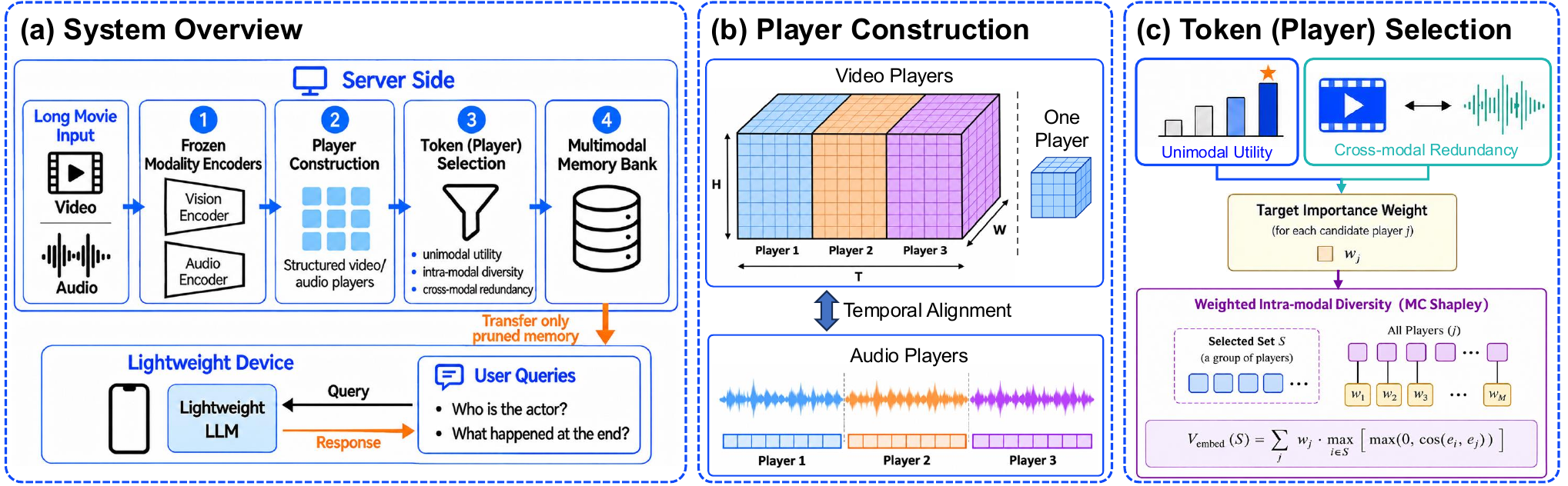}
  \caption{
    (a) The server encodes audio-visual content, constructs token groups as players, and performs query-agnostic selection to build a compact multimodal memory.
    The memory is transferred once to a lightweight device and reused to answer subsequent questions locally.
    Selection considers unimodal utility, intra-modal diversity, and cross-modal redundancy.
    (b) Video players are formed from spatio-temporal token blocks, while audio players group tokens aligned with the corresponding temporal intervals.
    Colors illustrate example groups in each modality.
    (c) We estimate each player’s Shapley value by Monte Carlo sampling of player permutations using a weighted coverage function that captures intra-modal diversity through cosine similarity of embeddings (i.e., $e$). 
    The weights incorporate unimodal utility, derived from token-embedding L2 norms as a proxy for semantic informativeness, and cross-modal redundancy, measured by cosine similarity between players from different modalities. 
    Players with higher estimated Shapley values are retained, while lower-scoring players are pruned.
  }
  \label{fig:overview}
\end{figure*}

\subsection{Multimodal Large Language Models for Video Learning and Understanding}
Multimodal large language models (MLLMs) extend language-based interaction to video by integrating visual, temporal, and auditory evidence into a language modeling framework.
A central challenge is to bridge pretrained perceptual encoders and language models while preserving the information required for video captioning, question answering, and event understanding.
Video-LLaMA~\cite{videollama} introduces modality-specific query transformers to connect visual and audio representations with a language model, enabling instruction-following video understanding grounded in both modalities.
Video-LLaVA~\cite{videollava} aligns image and video representations before projection into the language space, supporting joint learning from image and video inputs.
To better capture temporal dynamics, VideoLLaMA2~\cite{videollama2} introduces a spatio-temporal convolutional connector and incorporates an audio branch for audio-visual understanding.
Qwen2.5-Omni~\cite{qwen25} further supports unified text, image, audio, and video understanding, with temporally aligned audio-visual representations for processing multimodal streams.
More recently, video-SALMONN-2~\cite{videosalmonn2} improves audio-visual captioning through multi-round direct preference optimization and uses the resulting high-quality generated captions for supervised fine-tuning, transferring captioning improvements to video question answering.
Despite their strong performance, these models can incur substantial inference latency and energy consumption when processing long audio-visual sequences.
These costs limit their ability to support rich multimodal experiences on mobile and wearable devices with constrained computing and battery budgets.

\subsection{Token Pruning for Multimodal Large Language Models}
Token pruning and merging reduce multimodal inference costs by removing redundant representations.
For image-based MLLMs, existing methods select or compress visual tokens using criteria such as attention, semantic relevance, and diversity~\cite{sparsevlm,divprune,visionzip,vispruner,representationshift,ifprune}.
For video understanding, methods additionally exploit spatial and temporal redundancy~\cite{prunevid,dycoke,earlytom,unified_spatiotemporal_compression,streamingtom}.
Some further adapt token selection to a given question or the decoding process: PruneVid~\cite{prunevid} combines spatiotemporal merging with question-guided pruning, while DyCoke~\cite{dycoke} performs temporal compression and dynamic KV-cache reduction.
Such question- or decoder-dependent stages cannot be directly applied when a fixed memory must be constructed before future questions are known.
Beyond visual tokens, OmniZip~\cite{omnizip} compresses audio-visual inputs by using audio retention to guide video token allocation and cross-modal similarity to consolidate audio anchors.
We examine a different role for cross-modal similarity: it can indicate duplicated evidence and therefore a reduced benefit from retaining both representations.
Our method downweights cross-modal redundancy while preserving modality-specific utility and unimodal coverage.
This encourages the transferred memory to retain complementary evidence across audio and video within a limited token budget.

\subsection{Coverage Objectives and Cooperative Valuation}
Submodular objectives provide a natural language for representative subset selection because they model diminishing returns.
Prior work~\cite{NIPS2014_41321d69} on image collection summarization uses functions capturing coverage and diversity.
The Shapley value allocates a game's total value according to each player's expected marginal contribution~\cite{Shapley1953}.
Data Shapley applies this perspective to training-data valuation~\cite{ghorbani2019data}.
We instead treat audio-visual token groups as players and evaluate coalitions using a frozen embedding-space surrogate.
This avoids retraining or invoking the language decoder for coalition evaluation.
The role of the Shapley value here is to distribute representational coverage credit among groups, rather than to explain a particular answer or directly measure downstream answer accuracy.
\section{Methodology}
\label{sec:method}

Our method, as illustrated in Fig.~\ref{fig:overview}, constructs a compact, reusable multimodal memory for server-device inference.
The entire framework is training-free, with all parameters of the audio-visual encoders and the LLM kept frozen.
The server first encodes the input video and audio using frozen modality encoders and organizes their token embeddings into structured players (see Sect.~\ref{sec:groups}).
Next, we estimate the Shapley value of each player, using an embedding coverage objective that incorporates unimodal utility and cross-modal redundancy.
We select players under the allocated modality budgets and retain their original token embeddings as the memory.
The memory is transferred once to the device, where a lightweight decoder answers subsequent questions using the same stored representation.
This design reduces the multimodal context transmitted and processed locally while keeping user queries on the device.

\subsection{Problem Setup and the Memory Interface}
\label{sec:setup}

Our objective is to construct a useful memory under a limited token budget, which specifies the proportion of audio and video tokens to retain after pruning.
Let $x=(x^v,x^a)$ denote a media item with visual and audio streams.
We generate pre-LLM token sequences ($H$) using frozen modality encoders $F$, together with any existing fixed adapters at the chosen interface:
\begin{equation}
  H^m = F_m(x^m)
      = \{h^m_\ell\}_{\ell=1}^{N_m},
  \qquad m\in\{v,a\},
  \label{eq:encoding}
\end{equation}
where $N_m$ is the number of tokens in modality $m$.

Given modality budgets $k_v$ and $k_a$, we use a selector $\mathcal{P}$ to construct a compact memory from these token embeddings without access to user questions or intermediate LLM computations:
\begin{equation}
  \begin{aligned}
  \M(x) &= \mathcal{P}(H^v,H^a; k_v,k_a),\\
  \hat y &= D_\theta(q,\M(x)).
  \end{aligned}
  \label{eq:interface}
\end{equation}
For a given media item and set of budgets, the selected token indices remain fixed across questions.
We transfer the memory once and reuse it on the device, where the decoder $D_\theta$ answers subsequent questions.
This design reduces repeated transmission of multimodal context and keeps user questions on the device, helping protect user privacy.

\subsection{Structured Token Groups as Players}
\label{sec:groups}

Scoring tokens individually can overlook local structure and increase coalition evaluation costs.
Moreover, meaningful audio-visual events often span multiple spatial and temporal tokens.
We therefore group neighboring tokens into structured units, each treated as one player.

For video, we form spatio-temporal blocks using each token's source coordinates, accounting for the encoder's patch merging and output order.
Each block represents a spatial region over a temporal interval.
For audio, we group tokens by their native timestamps into windows aligned with the visual temporal blocks.

For modality $m\in\{v,a\}$, let $N_m$ denote the number of tokens and $N'_m$ the number of groups.
Each token belongs to exactly one group, and the token indices in group $i$ are denoted by $G_i^m$:
\begin{equation}
  \bigcup_{i=1}^{N'_m} G_i^m = \{1,\ldots,N_m\},
  \qquad
  G_i^m \cap G_j^m = \varnothing \quad \text{for } i\neq j.
  \label{eq:partition}
\end{equation}
We write $\G_m=\{1,\ldots,N'_m\}$ for the set of player indices.
To score player $i$, we measure the additional coverage it provides when added to a subset of other players. 
This subset is called a coalition and serves as the reference set for computing the marginal contribution of player $i$.
We estimate each player's Shapley value by averaging its marginal contributions over sampled random permutations, then retain the highest-scoring players within each modality's budget.

For valuation, each player is represented by an embedding descriptor
\begin{equation}
  e_i^m = \Pool_m\bigl(\{h_\ell^m:\ell\in G_i^m\}\bigr).
  \label{eq:pool}
\end{equation}
$\Pool_m$ is a fixed group aggregation operator.
These descriptors are used only for scoring: selecting a group retains its \emph{original token embeddings}, rather than replacing the group with the pooled descriptor.
We next describe how to compute the Shapley value for each player.

\subsection{Monte Carlo Shapley Valuation}
\label{sec:shapley}


Having constructed the players, we assign each player a score that measures its contribution to representing the encoded media.
We define a separate cooperative game for each modality $m\in\{v, a\}$, in which players form coalitions whose collective value is determined by a value function.

\paragraph{Coalition value and individual contribution.}
For player $i$, the marginal contribution is the additional coverage obtained by including $i$ in coalition $S\subseteq\G_m\setminus\{i\}$:
\begin{equation}
  \Delta_m(i\mid S)
  =
  V_m(S\cup\{i\})-V_m(S).
  \label{eq:marginal}
\end{equation}

For each modality, we define a value function that measures how well a selected subset of players represents all candidate players.
Let $e_i$ denote the embedding of player $i$ and $S$ the selected subset:
\begin{equation}
\begin{aligned}
  V_{\mathrm{embed}}(S)
  &= \sum_j w_j \max_{i\in S}
     \left[\max\left(0,\cos(e_i,e_j)\right)\right], \\
  V_{\mathrm{embed}}(\varnothing) &= 0.
\end{aligned}
\label{eq:value}
\end{equation}
Here, $j$ ranges over all candidate players within the modality, and $w_j$ is the importance weight of target player $j$.
Each target is represented by its most similar player in $S$.
This coverage objective encourages \emph{unimodal diversity}: adding a player that covers new content yields a larger marginal gain, whereas adding a near-duplicate contributes little.
We describe the construction of $w_j$ in Sec.~\ref{sec:value}.

\paragraph{Monte Carlo Shapley estimation.}
The Shapley value measures each player's average marginal contribution over all possible player orderings.
Directly enumerating these orderings is computationally prohibitive as the number of players grows.
We therefore approximate the Shapley value using Monte Carlo permutation sampling~\cite{ghorbani2019data}.

For each modality, we independently sample $R$ uniformly random permutations of its players.
Starting from an empty set, we add players in each sampled order and record the increase in $V_{\mathrm{embed}}$ after each addition.
Let $P_i^{\pi_r}$ denote the set of players preceding player $i$ in permutation $\pi_r$.
Its marginal contribution is
\begin{equation}
  \Delta_i^{(r)}
  =
  V_{\mathrm{embed}}\bigl(P_i^{\pi_r}\cup\{i\}\bigr)
  -
  V_{\mathrm{embed}}\bigl(P_i^{\pi_r}\bigr).
  \label{eq:sampled_marginal}
\end{equation}
We average these contributions to obtain the estimated Shapley value:
\begin{equation}
  \widehat{\phi}_i
  =
  \frac{1}{R}\sum_{r=1}^{R}\Delta_i^{(r)}.
  \label{eq:estimate}
\end{equation}
This procedure evaluates each player against different subsets of other players without exhaustive enumeration.
The resulting scores are used to retain the highest-ranked players within each modality's budget.

\subsection{Bi-signal Value Function}
\label{sec:value}

We now specify the weights used in Eq.~\eqref{eq:value}.
The target weight combines unimodal utility and cross-modal redundancy calibration.
The target weight is
\begin{equation}
  w_j^m=u_j^m c_j^m.
  \label{eq:weights}
\end{equation}
We define these two factors below.

\paragraph{Unimodal utility.}
Since user queries are unavailable during memory construction, we use embedding magnitude as a query-agnostic proxy for content importance.
Specifically, we compute each player's utility as the average L2 norm of its token embeddings:
\begin{equation}
  u_j^m
  =
  \frac{1}{|G_j^m|}
  \sum_{t\in G_j^m}
  \|h_t^m\|_2,
  \label{eq:utility}
\end{equation}
where $h_t^m$ is the frozen embedding of token $t$ in modality $m$.
This proxy assigns greater weight to groups with larger average embedding magnitudes without requiring query relevance scores.
In implementation, we normalize the utility scores to a unit mean within each modality.

\begin{table*}[t]
\centering
\caption{
QA accuracy (\%) on three benchmarks using two backbones.
The pruning ratio denotes the percentage of removed multimodal tokens.
Full Tokens denotes the unpruned model.
The best results are marked in \textbf{bold}.
}
\label{tab:main_results}

\small
\renewcommand{\arraystretch}{0.7}
\setlength{\aboverulesep}{2pt}
\setlength{\belowrulesep}{2pt}

\begin{tabular*}{\textwidth}{
  @{\extracolsep{\fill}} l c c c c @{}
}
\toprule
Method & Pruning Ratio & Video-MME & Daily-Omni & Video-Holmes  \\
\midrule
\multicolumn{5}{c}{\textit{video-SALMONN 2+ (3B)}} \\
\midrule
\rowcolor{gray!20}
Full Tokens & 0\% & 64.59 & 65.07 & 43.06 \\

DivPrune~\cite{divprune}    & & 61.70 & 52.30 & 39.63 \\
VisionZip~\cite{visionzip}   & & 59.04 & 51.04 & 38.54 \\
OmniZip~\cite{omnizip}     & & 61.62 & 52.80 & 39.51 \\
Ours & \multirow{-4}{*}{80\%} & \textbf{63.92} & \textbf{65.66} & \textbf{41.10} \\
\midrule

DivPrune~\cite{divprune}    & & 58.59 & 49.29 & 38.43 \\
VisionZip~\cite{visionzip}   & & 57.07 & 49.04 & 38.81 \\
OmniZip~\cite{omnizip}     & & 59.96 & 48.71 & 38.98 \\
Ours & \multirow{-4}{*}{90\%} & \textbf{61.22} & \textbf{61.32} & \textbf{39.36} \\
\midrule

\multicolumn{5}{c}{\textit{video-SALMONN 2 Pro (4B)}} \\
\midrule
\rowcolor{gray!20}
Full Tokens & 0\% & 73.89 & 77.11 & 48.62 \\

DivPrune~\cite{divprune}    & & 67.67 & 57.64 & 44.69 \\
VisionZip~\cite{visionzip}   & & 65.00 & 59.06 & 43.49 \\
OmniZip~\cite{omnizip}     & & 64.41 & 64.49 & \textbf{45.56} \\
Ours & \multirow{-4}{*}{80\%} & \textbf{68.41} & \textbf{73.68} & \textbf{45.56} \\
\midrule

DivPrune~\cite{divprune}    & & 65.52 & 56.64 & 44.26 \\
VisionZip~\cite{visionzip}   & & 63.59 & 56.89 & 42.68 \\
OmniZip~\cite{omnizip}     & & 63.62 & 64.49 & 45.29 \\
Ours & \multirow{-4}{*}{90\%} & \textbf{66.78} & \textbf{70.84} & \textbf{45.34} \\
\bottomrule
\end{tabular*}
\end{table*}

\paragraph{Cross-modal redundancy calibration.}
Audio and video may convey similar information about the same event.
The token embeddings within each player have already been projected into a shared embedding space.
We measure this overlap using cosine similarity between temporally aligned players from the two modalities:
\begin{equation}
  r_{j\ell}^{m,\bar m}
  =
  \cos(e_j^m,e_\ell^{\bar m}).
  \label{eq:cross_similarity}
\end{equation}
Higher similarity indicates greater potential redundancy between the two players.
Let $\mathcal{A}_m(j)$ contain the indices of players in the other modality that are temporally aligned with player $j$.
We define its redundancy score as the largest nonnegative similarity:
\begin{equation}
  d_j^m
  =
  \max_{\ell\in\mathcal{A}_m(j)}
  \max\{0,r_{j\ell}^{m,\bar m}\}.
  \label{eq:redundancy}
\end{equation}
If no aligned player is available, we set $d_j^m=0$.
We normalize these scores by their mean within each modality:
\begin{equation}
  \widetilde d_j^m
  =
  \frac{d_j^m}{\bar d_m},
  \qquad
  \bar d_m
  =
  \frac{1}{N'_m}
  \sum_{p\in\G_m}d_p^m.
  \label{eq:redundancy_normalization}
\end{equation}
When $\bar d_m$ is below the numerical tolerance, we leave the redundancy scores unchanged.
The calibration weight is
\begin{equation}
  c_j^m
  =
  \frac{1}{1+\lambda_{\mathrm{cross}}\widetilde d_j^m},
  \label{eq:calibration}
\end{equation}
where $\lambda_{\mathrm{cross}}$ controls the penalty strength and is set to $1$ in our experiments.
Players with greater relative cross-modal redundancy receive smaller calibration weights.
This downweights overlapping evidence and encourages the compact memory to preserve complementary information.

Before Shapley estimation, we normalize the resulting target weights to unit mean within each modality.
This common scaling does not change the within-modality ranking of Shapley scores.

\subsection{Dynamic Budget Allocation and Memory Construction}
\label{sec:budget}

A global pruning ratio determines the total number of retained tokens but leaves their allocation between audio and video unspecified.
The relative importance of audio and visual content can vary across media.
We therefore allocate the shared budget using the query-agnostic utility scores defined in Sec.~\ref{sec:value}.

\paragraph{Modality budget allocation}
Let $\mathcal{T}$ denote the set of modalities and $N_m$ the original token count of modality $m$.
Given a global pruning ratio $\rho$, the total retained-token budget is
\begin{equation}
  K=\left\lfloor(1-\rho)\sum_{m\in\mathcal{T}}N_m\right\rfloor.
  \label{eq:total_budget}
\end{equation}
For each modality, we aggregate the utilities of its players:
\begin{equation}
  U_m=\sum_{j\in\G_m}u_j^m,
  \qquad m\in\mathcal{T}.
  \label{eq:modality_utility}
\end{equation}
We allocate the shared budget according to these aggregate utilities:
\begin{equation}
  \widetilde{k}_m
  =
  K\frac{U_m^\alpha}
  {\sum_{\ell\in\mathcal{T}}U_\ell^\alpha},
  \label{eq:budget_allocation}
\end{equation}
where $\alpha>0$ controls the allocation preference for modalities with higher utility.
We round the allocations to integer budgets $k_m$ while preserving $\sum_{m\in\mathcal{T}}k_m=K$.
The resulting modality-specific pruning ratios are
\begin{equation}
  \rho_m=1-\frac{k_m}{N_m}.
  \label{eq:modality_pruning}
\end{equation}
These ratios adapt to the encoded media without requiring user queries.
In our audio-visual setting, $\mathcal{T}=\{v,a\}$.

\paragraph{Player selection and memory construction.}
Utility determines the budget assigned to each modality, while Shapley scores determine which players to retain within that budget.
We rank players by their estimated Shapley values and select groups subject to
\begin{equation}
  \sum_{i\in S_m}|G_i^m|\leq k_m.
  \label{eq:selection_budget}
\end{equation}
The resulting audio-visual memory is transferred once to the device and reused for subsequent local questions.
\section{Experiments}
\label{sec:experiments}

\subsection{Experimental Setup}
\label{sec:experimental_setup}

\subsubsection{Backbones}
We select models with relatively lightweight language decoders to accommodate the memory and computational constraints of user devices in our server--device setting.
We therefore evaluate our method on two audio-visual language models~\cite{videosalmonn2}: video-SALMONN 2+ (3B) and video-SALMONN 2 Pro (4B).
The former builds on Qwen2.5-VL, while the latter adopts Qwen3-VL.
Both models support joint audio-visual understanding and provide pretrained modality encoders and a language decoder.
We apply token selection to their encoded multimodal representations before language decoding, without updating model parameters.

\subsubsection{Benchmarks}
We evaluate on Video-MME, Video-Holmes, and Daily-Omni.
\textbf{Video-MME}~\cite{videomme} covers diverse video domains and durations, with 900 videos and 2,700 manually annotated question--answer pairs.
Its short-, medium-, and long-duration subsets allow evaluation across different temporal scales.
\textbf{Video-Holmes}~\cite{videoholmes} focuses on complex video reasoning, with 1,837 questions spanning seven tasks over 270 suspense short films.
It requires models to locate and connect visual clues across different video segments.
\textbf{Daily-Omni}~\cite{dailyomni} contains 684 real-world videos and 1,197 multiple-choice questions across six task families.
Its questions emphasize joint audio-visual reasoning and temporal alignment across modalities.
Together, these benchmarks assess general video understanding, reasoning over distributed evidence, and cross-modal information integration.



\subsubsection{Baselines}
We compare against three recent token compression methods published in 2025--2026: DivPrune~\cite{divprune}, VisionZip~\cite{visionzip}, and OmniZip~\cite{omnizip}.
We adapt their publicly available implementations to our backbones, focusing on methods that can construct compact representations before the user question is available.
Methods requiring query-conditioned selection~\cite{flexselect} are outside this comparison, as they cannot directly support our query-agnostic, one-time-transfer setting.

\subsection{Main Results}
\label{sec:main_results}

Table~\ref{tab:main_results} presents the main results in three benchmarks and two backbones, with unpruned models included as a baseline of the upper-bound.
Our method outperforms existing approaches in most evaluated settings.
For example, on DailyOmni at a pruning ratio of $90$\% with video-SALMONN 2+ (3B), our method surpasses DivPrune, VisionZip, and OmniZip by $12.03$, $12.28$, and $12.61$ percentage points, respectively.
These results demonstrate the effectiveness of our method in preserving task-relevant information under limited token budgets, even at high pruning ratios.

\subsection{Visualization}

\begin{figure}[t]
  \centering
  \includegraphics[width=0.42\textwidth]{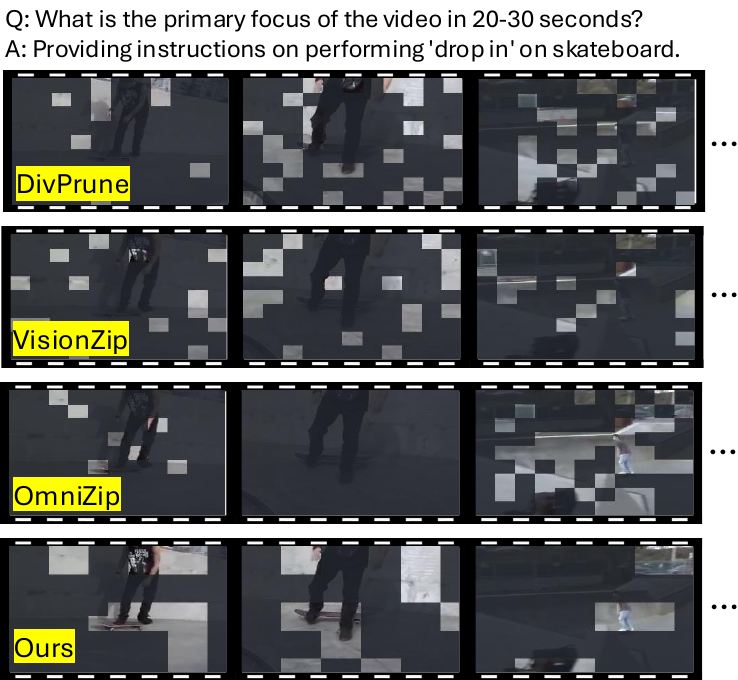}
  \caption{
    Each row shows the same sampled frames for a different method, with retained regions highlighted and pruned regions darkened.
    Our method better preserves key objects and complementary action-related cues across the displayed frames, including the skateboard, foot placement, and the demonstrator's body.
    ``Q'' and ``A'' denote the question about the video content and its corresponding ground-truth answer, respectively.
  }
  \label{fig:vis}
\end{figure}

To further illustrate the effectiveness of our method, we provide a qualitative comparison with DivPrune, VisionZip, and OmniZip across multiple video frames in Figure~\ref{fig:vis}. 
All methods are evaluated on the Daily-Omni dataset using video-SALMONN 2+ 3B with an identical pruning ratio of 80\%. 
In this example, DivPrune and VisionZip retain scattered visual regions while discarding much of the content associated with the skateboard and the demonstrator. 
OmniZip preserves some relevant regions, but their coverage varies substantially across the displayed frames, with the second frame almost entirely pruned. 
In contrast, our method retains the skateboard and the demonstrator's feet in the first two frames, preserving cues about foot placement relative to the board, while retaining a region covering the demonstrator's body in the later frame. 
These complementary cues across different moments help characterize the skateboarding instruction depicted in the video.  
This visualization offers qualitative evidence that our method preserves semantically informative content across both spatial regions and temporal samples under a constrained token budget.

\subsection{Ablation Studies}
\label{sec:ablation}

\paragraph{Effect of Monte Carlo Sampling.}
We investigate the sensitivity of our method to the number of Monte Carlo samples used for estimating Shapley values.
As shown in Table~\ref{tab:ablation}, increasing the sampling budget from 16 to 64 improves accuracy from 60.07\% to 61.32\%.
This improvement is consistent with the reduced estimation uncertainty expected from a larger sampling budget, which can lead to more reliable estimates of token contributions.
Interestingly, further increasing the number of samples to 256 slightly decreases accuracy to 60.57\%.
This may be because more accurate Shapley value estimation does not necessarily lead to better token selection for downstream question answering, as our value function serves as a proxy for task-specific utility. Overall, 64 samples achieve the best empirical performance among the evaluated settings while maintaining lower computational overhead than 256 samples.

\paragraph{Effect of Value Function Components.}
We further evaluate the contributions of two proposed weighting items in our value function.
Removing cross-modal redundancy modeling reduces accuracy from 61.32\% to 60.99\%, suggesting that downweighting redundant information shared between audio and video helps preserve more complementary evidence within a limited memory budget.
Similarly, removing feature utility weighting decreases accuracy to 61.24\%, indicating that individual feature informativeness provides an additional signal for token selection.
These results support the effectiveness of combining complementary weighting factors within our value function to construct compact yet informative audio-visual memory.

\begin{table}[t]
    \centering
    \caption{
    Ablation studies using video-SALMONN 2+ (3B) at a 90\% pruning ratio. 
    We investigate the number of Monte Carlo samples and the contributions of different value function components. The best results are highlighted in \textbf{bold}.}
    \label{tab:ablation}

    \begingroup
    \fontsize{8pt}{8pt}\selectfont  
    \renewcommand{\arraystretch}{0.8} 
    \setlength{\tabcolsep}{4pt}         
    \setlength{\aboverulesep}{0.2ex}    
    \setlength{\belowrulesep}{0.2ex}    

    \begin{adjustbox}{max width=\linewidth}
        \begin{tabular}{lccc}
            \toprule
            \multicolumn{4}{c}{\textbf{Monte Carlo Sampling}} \\
            \midrule
            \# Samples & 16 & 64 & 256 \\
            Accuracy (\%) $\uparrow$ & 60.07 & \textbf{61.32} & 60.57 \\
            \midrule
            \multicolumn{4}{c}{\textbf{Value Function Components}} \\
            \midrule
            Configuration & Original & w/o Cross-Mod. & w/o Feature Util. \\
            Accuracy (\%) $\uparrow$ & \textbf{61.32} & 60.99 & 61.24 \\
            \bottomrule
        \end{tabular}
    \end{adjustbox}

    \endgroup
\end{table}

\subsection{Efficiency Analysis}
\label{sec:efficiency}

In this section, we evaluate the efficiency of different methods on the DailyOmni dataset using video-SALMONN 2+ (3B) at an 80\% pruning ratio.

\paragraph{Computational Efficiency.}
Since we perform audio-visual token pruning before the LLM, we evaluate the computational efficiency of on-device LLM inference in terms of prefill and decoding latency. As shown in Table~\ref{tab:efficiency}, our method achieves the lowest prefill latency of 317.43 ms, delivering an 8.7$\times$ speedup over the unpruned baseline (2755.23 ms). It also achieves the lowest decoding latency of 50.71 ms, compared with 91.26--97.32 ms for existing pruning methods, yielding up to a 1.9$\times$ speedup. These improvements, together with our strong QA accuracy, demonstrate the effectiveness of our compact audio-visual memory in enabling efficient on-device multimodal understanding.

\paragraph{Memory Efficiency.}
Our method substantially reduces the memory footprint of retained input embeddings from 182.33 MiB to 36.90 MiB, achieving a 79.8\% reduction over the unpruned baseline.
Notably, the differences in memory footprint between our method and existing pruning approaches are negligible, while our method achieves stronger downstream QA performance (Sec.~\ref{sec:main_results}) by better preserving task-relevant information. 
Combined with its superior inference efficiency on the LLM side, as well as consistently competitive, often superior QA accuracy, these results highlight the effectiveness of our approach in balancing computational efficiency and task performance.

\begin{table}[t]
    \centering
    \caption{We report the average LLM prefill and decoding latency, together with the memory of retained input embeddings, assuming a hidden dimension of 2048 and BF16 precision. The best results are highlighted in \textbf{bold}.}
    \label{tab:efficiency}

    \begingroup
    \fontsize{8pt}{9pt}\selectfont  
    \renewcommand{\arraystretch}{0.80} 
    \setlength{\tabcolsep}{8pt}        
    \setlength{\aboverulesep}{0.2ex}   
    \setlength{\belowrulesep}{0.2ex}   

    \begin{adjustbox}{max width=\linewidth}
        \begin{tabular}{lccc}
            \toprule
            \textbf{Method} & \textbf{Prefill (ms)} $\downarrow$ & \textbf{Decode (ms)} $\downarrow$ & \textbf{Memory (MiB)} $\downarrow$ \\
            \midrule
            Unpruned & 2755.23 & 112.90 & 182.43 \\
            \midrule
            DivPrune~\cite{divprune} & 385.33 & 97.32 & \textbf{36.80} \\
            VisionZip~\cite{visionzip} & 318.02 & 91.26 & 36.85 \\
            OmniZip~\cite{omnizip} & 386.48 & 96.72 & 36.85 \\
            \textbf{Ours} & \textbf{317.43} & \textbf{50.71} & 36.90 \\
            \bottomrule
        \end{tabular}
    \end{adjustbox}

    \endgroup
\end{table}

\paragraph{Network-Aware Amortized Latency.}
Unlike conventional per-query token pruning, our framework targets a server--device scenario in which compact, query-agnostic audio-visual memory is constructed once and reused to answer multiple questions locally.
Since memory construction is independent of user queries, the encoded and pruned representations can be precomputed and cached on the server before any question arrives, following the offline memory preparation paradigm adopted in prior video understanding systems~\cite{di2025rekv, Xiao_2026_CVPR}.
We therefore exclude server-side encoding and pruning costs from the online latency analysis and focus on memory transmission and device-side inference.

\begin{figure}[t]
  \centering
  \includegraphics[width=0.48\textwidth]{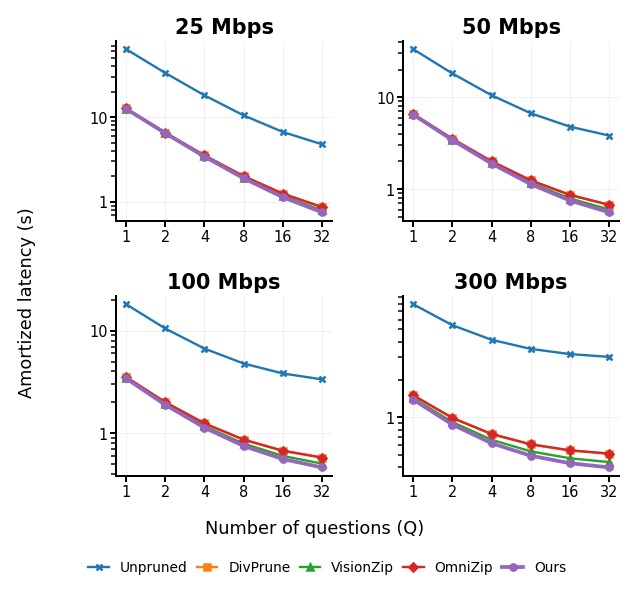}
  \caption{
  Network-aware amortized latency under different assumed bandwidths (25, 50, 100, and 300 Mbps). 
  Server-side encoding and pruning are excluded under the assumption that query-agnostic memory is precomputed and cached before online question answering. 
  Transfer latency is analytically estimated from the memory payload size, while LLM inference costs are obtained from measured stage-wise timings. 
  Lower is better.
  }
  \label{fig:amortized_latency}
\end{figure}

For a video associated with $Q$ questions and an effective network bandwidth $B$, we define the average amortized latency per question as
\begin{equation}
\begin{aligned}
T_{\mathrm{avg}}(Q,B)
&= \frac{T_{\mathrm{transfer}}(B)}{Q}
+ T_{\mathrm{prefill}} + T_{\mathrm{decode}}, \\
T_{\mathrm{transfer}}(B)
&= \frac{M \times 2^{20} \times 8}{B \times 10^6},
\end{aligned}
\label{eq:amortized_latency}
\end{equation}
where $M$ denotes the estimated memory payload size in MiB and $B$ is measured in Mbps.
The transfer latency follows the standard data-size-over-bandwidth model~\cite{kurose2021networking}.
We evaluate $Q \in \{1,2,4,8,16,32\}$ under bandwidths of 25, 50, 100, and 300 Mbps.
As illustrated in Fig.~\ref{fig:amortized_latency}, our method consistently achieves the lowest estimated amortized latency among the compared methods under this offline-preprocessing assumption.
Moreover, its relative latency advantage over existing pruning methods becomes increasingly pronounced as more questions reuse the same memory.
For example, compared with VisionZip, our latency reduction increases from 0.22\% to 5.47\% at 25 Mbps as $Q$ increases from 1 to 32.
These results highlight our reusable memory’s benefits for repeated on-device QA, complementing the accuracy and efficiency gains reported above.

\section{Conclusion}
\label{sec:conclusion}



{\looseness=-1
We presented a training-free, query-agnostic framework that constructs compact audio-visual memories for one-time transfer and on-device reuse.
Across evaluated models and benchmarks, our method achieves higher QA accuracy than competing token compression methods by selecting informative and complementary evidence.
Future work can explore more multimodal large language models, detector-based player construction, and Shapley scores as soft supervision for a learnable player selector.
\par}


{
    \small
    \bibliographystyle{ieeenat_fullname}
    \bibliography{main}
}


\end{document}